# Multiple decision trees

*Suk Wah Kwok†*
*Chris Carter*

Basser Department of Computer Science
Madsen Building, F09, University of Sydney, N.S.W. 2006, Australia

## ABSTRACT

This paper describes experiments, on two domains, to investigate the effect of averaging over predictions of multiple decision trees, instead of using a single tree. Other authors have pointed out theoretical and commonsense reasons for preferring the multiple tree approach. Ideally, we would like to consider predictions from all trees, weighted by their probability. However, there is a vast number of different trees, and it is difficult to estimate the probability of each tree. We sidestep the estimation problem by using a modified version of the ID3 algorithm to build good trees, and average over only these trees. Our results are encouraging. For each domain, we managed to produce a small number of good trees. We find that it is best to average across sets of trees with different structure; this usually gives better performance than any of the constituent trees, including the ID3 tree.

## 1. Introduction

A common goal in machine learning is to make predictions based on previous data. In this paper we consider classification tasks – where predictions are restricted to assigning data points to one of a small number of mutually exclusive and exhaustive classes.

Many classification systems use the data to select one model – from a given class of models – and then use this model to make predictions. Using Self and Cheeseman's terminology (Self and Cheeseman, 1987) we call this method *abduction*. Many authors have pointed out that this method is not optimal (Howard, 1970; Self and Cheeseman, 1987). Self and Cheeseman give an example involving a military commander who is unsure of the location of his enemy. Abduction corresponds to chosing the most likely location, and then proceeding as if it were a fact.

Ideally, we would like to use predictions from all possible models, weighted by the probability of each model. We call this method *transduction* (Self and Cheeseman, 1987) Using the military example, we should consider all possible locations of the enemy, not just the most likely. The difference between these methods is greatest when there is a small amount of data available. In this case, there is little basis for preferring any one model over the others, so abduction may give misleading results.

When using transduction, however, it may be computationally infeasible to consider all possible models, and it may be difficult to estimate the probability of each model. The practical result is that we are restricted to models and probabilities that facilitate the calculations.

---

† Currently a knowledge engineer at the Australian Mutual Provident Society (AMP) in Sydney.
  e-mail sukwah@basser.cs.su.oz

This paper describes a hybrid approach proposed by Buntine (Buntine, 1987). We use *decision trees*, a class of models that has performed well in classification problems (Quinlan, Compton, Horn and Lazarus, 1987), and average over a small number of models each with a high probability. We hope to get some of the benefits of transduction, whilst retaining the complex models possible with abduction. Our results are encouraging; we find improved performance with a very small number of models on a variety of domains. We show that, for the small number of models we used, it is important that the models be as different as possible. Since we had trouble constructing many different models, this is a key area for future work.

## 2. Overview of ID3

The ID3 algorithm, and related algorithms, have been used with considerable success in classification tasks (Quinlan et. al., 1987; Breiman, Friedman, Olshen and Stone 1984; Michie, 1987). It is normally used to select a single model from data; this model is then used to predict the class of future data.

The data is described by a fixed set of *attributes*. The attributes are chosen to be predictive of the class. For example, if we were considering first year computer science students, and the class we wish to predict is whether the student passed or failed first year computer science, the following attributes might be appropriate.

**High school result** with numeric values.
**High school maths result** with numeric values.
**Previous programming experience** with values {yes and no}.
**Temperment** with values {patient and impatient}.

ID3 uses a *training set* of classified data to construct a decision tree. Figure 1 shows a decision tree for the student example. Each internal node of the tree contains a test on the value of an attribute, and each leaf of the tree is assigned a class.

```
high school result < 62.5%: fail (2)
high school result > 62.5%:
|       temperment = patient: pass (3)
|       temperment = impatient:
|       |       high school maths result < 82.5%: fail (2)
|       |       high school maths result > 82.5%: pass (1)
```

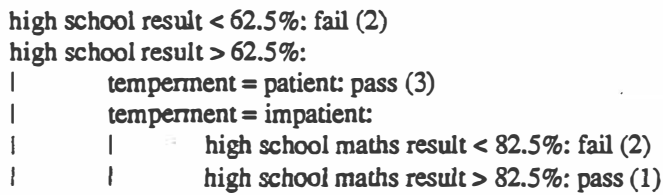

**Figure 1.** A decision tree

To construct this tree, ID3 uses a top-down approach, first chosing the test for the root of the tree, and then working downwards. Tests are chosen using a heuristic called the maximum *information-gain* (Quinlan, 1986), which tries to build a simple tree that fits the training set. The algorithm terminates when the tree correctly classifies the training set.

Once a tree has been built, it can be *pruned* by collapsing subtrees into leaves. This has been shown to generally improve performance (Quinlan, 1987a). After pruning some leaves may have training set objects of different classes. The proportion in each class can be used as an estimate of the probability of each class. We call this a *class probability tree* (Breiman, et. al., 1984; Carter and Catlett, 1987; Quinlan, 1987b).

## 3. Background theory

The motivation for our work comes from a Bayesian analysis of decision tree methods by Buntine (Buntine, 1987). This analysis shows that it is difficult to calculate the posterior probability of decision tree models. The difficulty lies in calculating the prior probability of the trees. Thus a strict application of the transduction approach does not seem possible. Buntine

214

proposes a hybrid approach, which we adopt, where a few trees of high posterior probability are used to get some of the averaging effect of transduction.

**Bayesian analysis**

Buntine considers the case when there are two classes, (positive and negative) and a finite number of possible data points (that is, all attributes take discrete values). Let $C$ be the number of different data points, and $\phi_i$ be the probability of type $i$ data point having positive class, for $i = 1$ to $C$. We call the vector $\Phi = (\phi_1, \phi_2, ..., \phi_C)$ a classification rule.

The Bayesian approach is to work with the distribution, $p(\Phi)$, and analyse how this distribution is updated by the training set, $\mathbf{x}$. Thus, we wish to calculate

$$p(\Phi \mid \mathbf{x})$$

using Bayes' Theorem

$$p(\Phi \mid \mathbf{x}) = \frac{p(\mathbf{x} \mid \Phi) p(\Phi)}{p(\mathbf{x})}$$

where $p(\mathbf{x})$ is a normalising constant. Buntine shows that, if there are $n_i$ type i data points in the training set, $r_i$ of which have positive class, then

$$p(\mathbf{x} \mid \Phi) = \Pi_{i=1,...,C} \phi_i^{r_i} \phi_i^{n_i - r_i}$$

so the updating formula for classification rules is straightforward.

However, Buntine points out difficulties in the choice of the prior $p(\Phi)$. He argues that the success of ID3 with small training sets (a fraction of the number of data point types) means that strong correlations between data must be justified a priori. He also points out that the prior must incorporate Occam's Razor because of the observation that simple trees are superior, and because simple trees are easier to understand, so they are a more effective representation of domain knowledge.

**A simplified approach**

These difficulties lead us to follow Buntine's suggestion and abandon (at least for the time being) calculating the posterior probability of each classification rule. Instead, we generate several rules of high posterior probability and average across them.

We're interested in two main areas: the distribution of classification rules, $p(\Phi \mid \mathbf{x})$, and the effect of combining a small number of classification rules with high posterior probability.

The shape of the distribution of classification rules, $p(\Phi \mid \mathbf{x})$, is of particular interest. If it is fairly flat (so there are many rules of equal posterior probability) then the abduction approach of choosing one rule may give poor results. However, if it has a sharp peak, (a few rules dominate) then the abduction approach is probably adequate.

Regarding the combination of classification rules, we wish to know whether the performance of the combination is better than the performance of single rules. We would also like to know how the number of rules affects the performance of the combination.

## 4. Experiments

To investigate the areas mentioned above, we performed a number of experiments using decision trees built by a interactive version of ID3 (Kwok, 1987). We chose ID3, and decision trees, because of their successful track record in classification tasks. This track record suggests that the ID3 decision trees have reasonably high posterior probability, though possibly not optimal.

215

These experiments address the following questions:

- Is the ID3 tree special? That is, can we build other simple trees with competitive performance, and if so, how many?
- Does averaging across a group of good trees improve accuracy?
- What happens when we increase the number of trees?

### Building multiple trees

Because we weight the trees equally when calculating the average, we require a systematic method of building many trees of comparable accuracy to the original ID3 tree. Recall from Section 3 that ID3 generates a single tree in a top-down manner, at each node choosing the test with the highest information gain. We adopt an interactive version of the ID3 algorithm that allows the user to select the test. We still calculate the information gain for each test, and we present the user with a ranked list of tests for guidance, with the highest information gain test as the default.

We find that, if we use a test with an information gain that is close to the maximum, the resulting tree does not perform significantly worse. In fact, some trees perform better.

In our experiments, we only try to override the default test at the top tree levels. This is because the choice of test at the top tree levels affects all the tests that appear below it; by doing this we hope to obtain trees that are substantially different from each other.

### Combining multiple trees

We use each tree independently to classify data points and then average the results. There are several ways of doing this. First, individual trees can return a categorical result (decision trees) or probability estimates (class probability trees). In both cases, we can take the estimates and average them. This average can be changed to a categorical result, or left as a probability estimate.

To interpret the average categorically, we choose the class with the highest estimate. We call this the *voting* method. A result from Buntine (Buntine, 1987) shows that the voting method minimizes the expected error, given that we must return a categorical result. To evaluate performance, we calculate the percentage error over a test set of data.

We call leaving the average as a probability estimate the *class probability* method. We evaluate the performance of this method by calculating the Half-Brier score (one-half the mean square error) (Brier, 1950) over a test set of data.

### 5. Results

We performed experiments using two domains, Weather and Student, selected from a set of ten candidate domains (Kwok, 1987). We selected these particular domains because ID3 gives moderate performance on them, and there are several good attributes to choose at the root of the tree. Reasonable performance is important because, first, if performance is too poor ID3 is not an appropriate method anyway, and, second, if the performance is too good it is difficult to detect any improvement.

### Weather data

The weather data (Allen, Bolam and Ciesielski 1986) contains daily meteorlogical observations described by 35 attributes including humidity, pressure pattern, wind direction, cloud amount and temperature. The task is to decide whether it will rain on the following day. There are 5516 observations: 3000 in the training set, and 2516 in the test set.

### Student data

The student data (Kwok, 1987) contains information on first year Computer Science students at the University of Sydney in 1983. Students are described by attributes such as High School mark, High School maths units and mark, faculty, sex, and answers to two questionaires on their background in computing, opinions on the computer science course and expectation of their results. The task is to determine the performance of a student in the computer science course. There are two classes: pass or fail. The data contains 446 observations: 250 in the training set, and 196 in the test set.

### Performance of individual trees

For each domain, a number of trees were built using the interactive method described in Section 4. We also pruned each tree. Table 1 gives the accuracy and size (measured by the number of nodes) of individual trees before and after pruning. Tree 0 is the ID3 tree (built uninterrupted).

|  | Weather | | | | Student | | | |
|---|---|---|---|---|---|---|---|---|
|  | unpruned | | pruned | | unpruned | | pruned | |
| tree | % error | size | % error | size | % error | size | % error | size |
| 0 | 31.48 | 915 | 29.61 | 523 | 27.04 | 160 | 24.49 | 23 |
| 1 | 32.71 | 909 | 30.29 | 507 | 22.96 | 162 | 24.49 | 31 |
| 2 | 30.37 | 887 | 28.97 | 479 | 25.00 | 159 | 21.94 | 18 |
| 3 | 32.51 | 913 | 30.09 | 523 | 26.02 | 149 | 21.94 | 39 |
| 4 | 32.43 | 917 | 30.52 | 501 | 25.00 | 162 | 23.47 | 34 |
| 5 | 30.05 | 887 | 28.50 | 509 | 29.08 | 164 | 20.41 | 27 |
| 6 | 33.74 | 889 | 31.20 | 539 | 27.55 | 154 | 24.49 | 23 |
| 7 | 32.23 | 927 | 29.77 | 515 |  |  |  |  |
| Average | 31.94 | 906 | 29.87 | 512 | 26.09 | 159 | 23.03 | 28 |

**TABLE 1.** Performance of individual trees

In both domains our tree-building method (Section 4) found a group of trees with competitive performance to the ID3 tree (8 for the weather data, 7 for the student data). The ID3 tree is not the best tree in any of the domains, though it is a good tree. (2 better trees for the weather data, 4 for the student data).

In the student domain we obtain a larger group of good trees than in the weather domain. (in the case of pruned trees, 4 have better performance than the pruned ID3 tree, with trees 2, 3 and 5 significantly better.) This may be expained by the fact that in the student domain there are 6 or 7 attributes that yield good information gain at the top few levels of the ID3 tree, whereas, in the weather domain, there are only 2 or 3 good attributes.

### Different combinations of trees

We used the weather data to experiment with different combinations of unpruned and pruned trees. The weather data has a limited number of good attributes at the root of the tree, so some of the 8 trees we built are fairly similar – either having common roots or common second tree level nodes. This allows us to compare the performance of combinations of similar trees to combinations of different trees.

We expect that combining similar trees will not result in much improvement in accuracy. An intuitive explanation for this is that the group of similar trees give a "biased vote" for a particular class.

We tried different combinations of three trees; the individual result from each tree is interpreted as a probability estimate, and combined using the voting method to return a categorical result (as described in Section 4). Table 2 confirms our prediction that combinations of different trees have better performance than combinations of similar trees (common roots or common second tree level nodes). This is the case for both unpruned and pruned trees.

|  | 3 fairly different trees | | | 3 trees with common second level node | | | 3 trees with common root | |
|---|---|---|---|---|---|---|---|---|
|  | % error | | | % error | | | % error | |
| comb | unpruned | pruned | comb | unpruned | pruned | comb | unpruned | pruned |
| 0 1 2 | 28.42 | 27.98 | 0 1 4 | 31.20 | 29.01 | 0 2 4 | 31.00 | 29.25 |
| 0 1 3 | 29.33 | 28.06 | 0 1 5 | 29.01 | 28.38 | 0 2 5 | 30.48 | 28.50 |
| 0 2 3 | 29.17 | 28.06 | 0 1 6 | 32.04 | 29.69 | 0 4 5 | 31.48 | 29.49 |
| 0 2 6 | 28.50 | 27.90 | 0 2 4 | 31.00 | 29.25 | 1 3 6 | 32.00 | 29.77 |
| 0 2 7 | 28.50 | 27.86 | 0 2 5 | 30.48 | 28.50 | 1 3 7 | 32.23 | 29.57 |
| 0 6 7 | 28.93 | 27.42 | 0 3 4 | 30.68 | 28.74 | 3 6 7 | 31.52 | 29.25 |
| Average | 28.81 | 27.88 |  | 30.74 | 28.93 |  | 31.45 | 29.31 |

TABLE 2. Different combinations of three pruned trees

**The effect of using more trees**

To investigate whether increasing the number of trees improves accuracy, we progressively increase the number of pruned and unpruned trees. For unpruned trees, since individual trees return a categorical result, odd numbers of trees were used to avoid ties. For each number of trees, the average percentage error over many different combinations is given in Table 3. The combinations were chosen so that the constituent trees are as different as possible. So some possible combinations were ignored because the constituent trees were too similar.

|  | Weather | | | | Student | | | |
|---|---|---|---|---|---|---|---|---|
|  | unpruned | | pruned | | unpruned | | pruned | |
| # trees | %error | # comb. | %error | # comb. | %error | # comb. | %error | # comb. |
| 1 | 31.94 | 8 | 29.87 | 8 | 26.09 | 7 | 23.03 | 7 |
| 2 |  |  | 30.23 | 27 |  |  | 21.04 | 21 |
| 3 | 28.62 | 23 | 27.52 | 23 | 23.92 | 35 | 19.49 | 35 |
| 4 |  |  | 27.33 | 9 |  |  | 19.48 | 35 |
| 5 | 28.33 | 9 | 27.2 | 9 | 24.05 | 21 | 19.00 | 21 |
| 6 |  |  | 26.96 | 5 |  |  | 18.80 | 7 |
| 7 | 28.66 | 6 | 27.25 | 6 | 24.49 | 1 | 19.90 | 1 |
| 8 |  |  | 27.19 | 1 |  |  |  |  |

TABLE 3. Percentage error for increasing number of trees

For the limited numbers considered, increasing the number of trees initially improves performance. After only a few trees, however, the error rate reaches an optimal point; after that, additional trees slightly degrade performance. The overall improvement in performance is significant. In the case of unpruned trees, using multiple trees gives an improvement of 3.6% for weather and 2.2% for student. As a comparison, Table 1 (averages row) shows that the well-established technique of pruning gives an improvement of 2.1% for weather and 3% for student. For pruned trees, using multiple trees gives a further average improvement of 2.9% for weather and 4.2% for student over single trees.

**Class probability method**

So far we have used the voting method to return a categorical result. We now investigate the effect of returning the average class probability estimate. We used the Half-Brier score as our measure of error. To give some feel for the Half-Brier score, if all probability estimates are 0 or 1, the Half-Brier score corresponds to the error rate. Figure 2 shows a similar pattern to Table 3 (using the voting method). The Half-Brier score (a measure of error) decreases as the number of

218

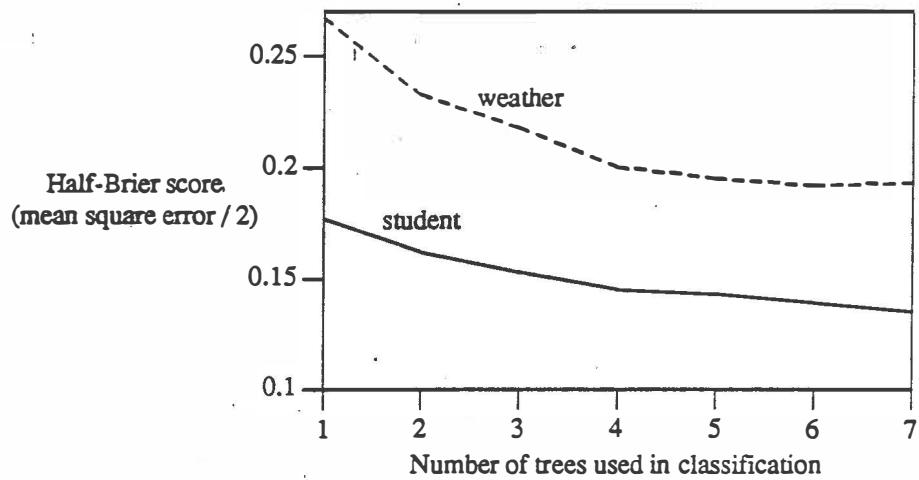

**Figure 2.** Error (Half-Brier score) against number of pruned trees

trees increase. The decrease is initially large and gradually slows down as more trees are added.

## 6. Conclusion

The purpose of our experiments was to see whether a limited form of transduction could improve the accuracy of classification methods using decision trees. Our limited form of transduction is to only use good trees, since we do not know how to accurately calculate the posterior probability of trees.

For each domain, our tree-building method produced a small number of trees of competitive size and performance to the ID3 tree. We found that it is best to average across sets of different trees; this usually gives better peformance than any of the constituent trees, including the ID3 tree. Unfortunately, our tree-building method did not produce sufficient numbers of different trees to allow a thorough investigation of the effect of increasing the number of trees.

We would like to find a tree-building algorithm that produces more good trees. If we cannot build sufficient good trees, we would like to investigate the effect of weighting the individual trees, however approximate are the weights.

## 7. Acknowledgement

We would like to thank Wray Buntine for his many useful suggestions on this work.